\documentclass[conference]{IEEEtran}
\IEEEoverridecommandlockouts
\usepackage{cite}
\usepackage{amsmath,amssymb,amsfonts}
\usepackage{algorithmic}
\usepackage{graphicx}
\usepackage{textcomp}
\usepackage{xcolor}

\usepackage{url}
\usepackage{graphicx}
\usepackage{epstopdf}

\usepackage{footnote}
\usepackage{balance}
\usepackage{amssymb}

\usepackage{amsmath}
\usepackage{multicol}
\usepackage{multirow}
\usepackage{anyfontsize}
\usepackage{array}
\usepackage{hhline}
\usepackage{float}
\usepackage{enumitem}
\usepackage{tabularx}

\def\BibTeX{{\rm B\kern-.05em{\sc i\kern-.025em b}\kern-.08em
    T\kern-.1667em\lower.7ex\hbox{E}\kern-.125emX}}

\begin{document}

\title{Detecting speaking persons in video}

\author{\IEEEauthorblockN{Hannes Fassold}
\IEEEauthorblockA{
\textit{JOANNEUM RESEARCH - DIGITAL}\\
Graz, Austria \\
hannes.fassold@joanneum.at}

}

\maketitle

\begin{abstract}
 We  present a novel method  for detecting speaking  persons  in  video, by extracting facial landmarks with a neural network and  analysing these landmarks statistically over time.
\end{abstract}


\section{Introduction}
Knowing which persons are actually speaking in a video at a certain time is an important semantic information which is useful in several application fields. We therefore present a novel robust method for detecting speaking persons in video which relies only on the visual information. It first extracts for all persons visible in the video their facial landmarks with a high-quality deep learning approach (see section \ref{sec:landmarks}). Afterwards, a statistical analysis of the landmark trajectories over time is employed to detect the temporal segments in which a certain person is speaking (see section \ref{sec:detector}).

\section{Facial landmark metadata extraction}
\label{sec:landmarks}
For the extraction of the facial landmark metadata, for each frame of the video sequence we first invoke the SF3D face detector from \cite{Zhang2017} in order to detect the ROIs in the frame which correspond to a face. For each detected face, we extract its facial landmarks with the 2D-FAN algorithm proposed in the work \cite{Bulat2017}. So for each face in each frame of the video, we retrieve a face descriptor containing the facial landmarks and some other data. In order to build a trajectory of the facial landmarks of one person over time, we have to group together all face descriptors belonging to the same person. For that, we currently employ a simple clustering strategy, based on the location of the extracted facial landmarks in the image.

\section{Detector algorithm}
\label{sec:detector}
\begin{figure}[t]
    \centering
    \includegraphics[width = 0.40\textwidth]{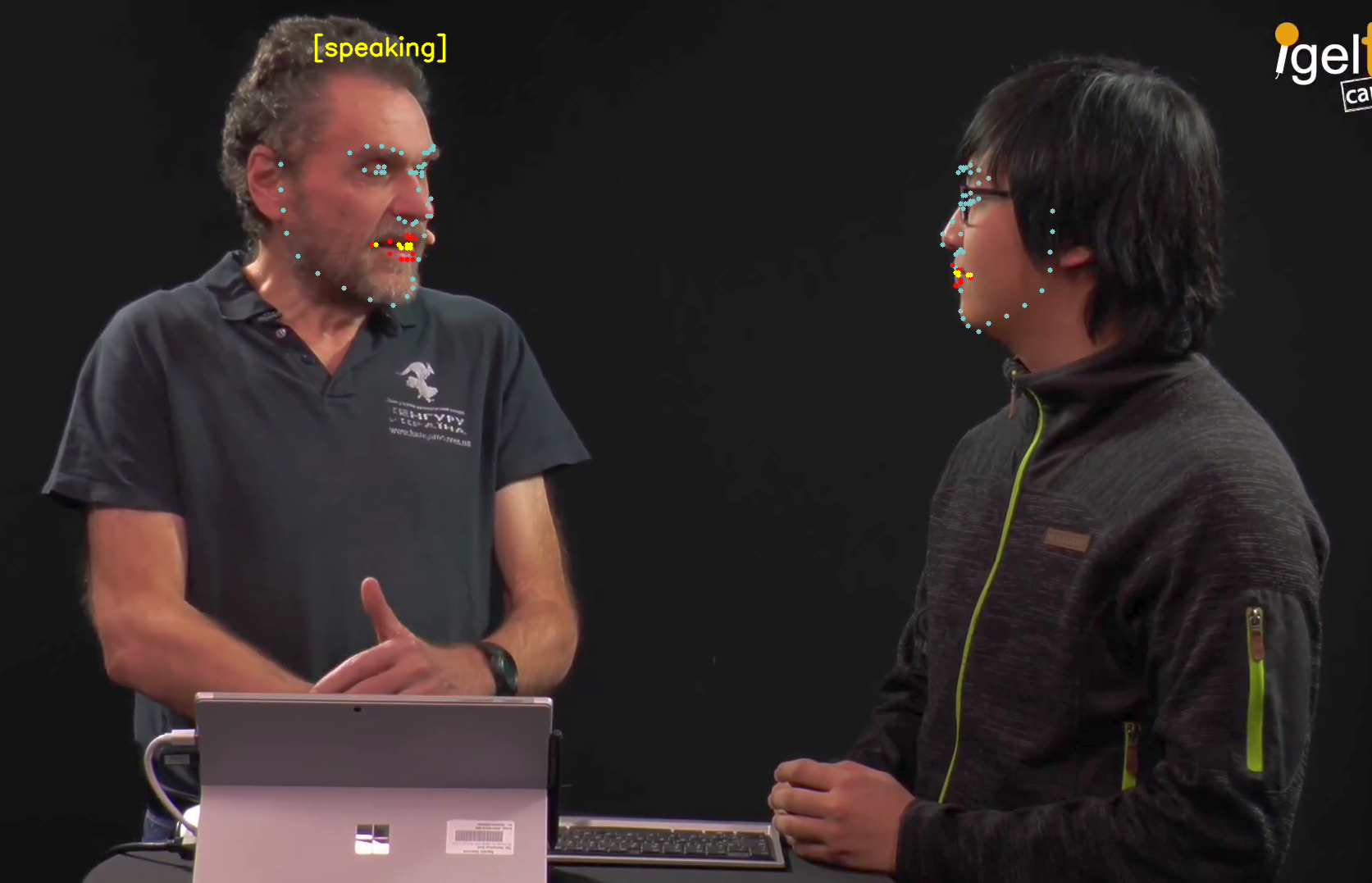}
    \caption{Speaking person detector with visualized facial landmarks. Image courtesy of IgelTV.}
    \label{fig:detection}
\end{figure}

For each person occurring in the video, we do now a statistical analysis of its facial landmark trajectory over time in order to infer whether that person is speaking or not and to detect in which time intervals the person does speak. When a person is speaking, he/she is repeatedly opening and closing the mouth. So we design our analysis method to detect this temporal pattern, based on the extracted landmarks for the mouth (visualized in yellow and red in Fig. \ref{fig:detection}). The basic idea is to measure the deviation in $y$-direction of the landmarks located at the lips and accumulate them over a few seconds, with a high deviation (significant lip movement) indicating that the person is speaking currently. As a preprocessing step, we have to \emph{normalize} the 68-dimensional landmark vectors, thereby making them invariant to their spatial position and the face height . For this, we shift each vector so that its coordinate origin $(0,0)$ is located at the middle of the mouth. Additionally, we scale each vector by the estimated face height (calculated from the top and bottom facial landmarks). We construct now a 1-D vector $t$ from the normalized $y$-coordinate of the facial landmark located at the top of the upper lip  for each frame. In the same way, we construct a 1-D vector $b$ from the respective facial landmark located at the bottom of the lower lip. In order to calculate the deviation of both vectors over an accumulation period of a few seconds, we calculate now significantly blurred versions of both vectors $\tilde{t}=smooth(t)$ and $\tilde{b}=smooth(b)$. The deviation vectors are now calculated as $d=abs(t-\tilde{t})$ and $e=abs(b-\tilde{b})$, and a combined deviation vector $c$ is calculated as the average of $d$ and $e$. We retrieve the time intervals where a speaking person is detected now by thresholding the combined deviation vector $c$, using a user-specified threshold $T$.



\section*{Acknowledgment}
This work was supported by European Union´s Horizon 2020 research and innovation programme under grant number 951911 - AI4Media.

\bibliographystyle{IEEEtran}
\bibliography{references}

\end{document}